\documentclass[11pt, a4paper, logo, copyright]{googledeepmind}

\usepackage{booktabs}
\usepackage{tabularx}
\usepackage{ragged2e} 

\usepackage[authoryear, sort&compress, round]{natbib}

\usepackage{amsmath,amsfonts,bm}

\def\eqref#1{equation~\ref{#1}}

\def\1{\bm{1}}

\DeclareMathAlphabet{\mathsfit}{\encodingdefault}{\sfdefault}{m}{sl}
\SetMathAlphabet{\mathsfit}{bold}{\encodingdefault}{\sfdefault}{bx}{n}

\usepackage{times}
\usepackage{latexsym}
\usepackage[T1]{fontenc}
\usepackage[utf8]{inputenc}
\usepackage{microtype}
\usepackage{inconsolata}
\usepackage{graphicx}
\usepackage{balance}       
\usepackage{graphics}      
\usepackage{hyperref}
\usepackage{color}
\usepackage{booktabs}
\usepackage{textcomp}
\usepackage{subcaption}
\usepackage{enumerate}
\usepackage{xcolor}
\usepackage{lipsum}
\usepackage{makecell}
\usepackage{multicol}
\usepackage{multirow}
\usepackage{array}
\usepackage{verbatimbox}
\usepackage{enumitem}
\usepackage{amsmath}
\usepackage{float}
\usepackage{stfloats}
\usepackage{graphicx}
\usepackage{amsthm}
\usepackage{listings}
\usepackage{caption} 
\usepackage[export]{adjustbox}
\usepackage{xspace}
\usepackage{epsfig}
\usepackage[linesnumbered]{algorithm2e}
\usepackage{algpseudocode}
\usepackage{tabularx}
\usepackage{arydshln}
\usepackage[bottom]{footmisc}
\usepackage{tcolorbox}
\usepackage{stackengine}
\usepackage{placeins}

\usepackage[normalem]{ulem}
\useunder{\uline}{\ul}{}
\tcbuselibrary{breakable}

\definecolor{E+F}{RGB}{	255, 99, 71}
\definecolor{B+F}{RGB}{255, 165, 0}
\definecolor{E+I}{RGB}{	173, 216, 230}
\definecolor{B+I}{RGB}{	30, 144, 255}
\definecolor{D}{RGB}{	60, 179, 113}

\definecolor{maroon}{cmyk}{0,0.87,0.68,0.32}

\usepackage{tikz}

\definecolor{darkgreen}{rgb}{0.0, 0.5, 0.0}
\definecolor{usercolor}{RGB}{200, 230, 250} 
\definecolor{coachcolor}{RGB}{180, 250, 180} 

\title{The Human–AI Hybrid Delphi Model: A Structured Framework for Context-Rich, Expert Consensus in Complex Domains}
\reportnumber{} 
\renewcommand{\today}{2025-07-25}

\author[1$\dagger$]{Cathy Speed}
\author[1$\ddagger$]{Ahmed A. Metwally}

\affil[1]{Google Research}
\affil[$\dagger$]{Work done while at Google Research}

\correspondingauthor{aametwally@google.com,}

\begin{abstract}

Expert consensus plays a critical role in domains where evidence is complex, conflicting, or insufficient for direct prescription. Traditional methods, such as Delphi studies, consensus conferences, and systematic guideline synthesis, offer structure but face limitations including high panel burden, interpretive oversimplification, and suppression of conditional nuance. These challenges are now exacerbated by information overload, fragmentation of the evidence base, and increasing reliance on publicly available sources that lack expert filtering. This study introduces and evaluates a Human–AI Hybrid Delphi (HAH-Delphi) framework designed to augment expert consensus development by integrating a generative AI model (Gemini 2.5 Pro), small panels of senior human experts, and structured facilitation. The HAH-Delphi was tested in three phases: retrospective replication, prospective comparison, and applied deployment in two applied domains (endurance training and resistance and mixed cardio/strength training). The AI replicated 95\% of published expert consensus conclusions in Phase I and showed 95\% directional agreement with senior human experts in Phase II, though it lacked experiential and pragmatic nuance. In Phase III, compact panels of six senior experts achieved >90\% consensus coverage and reached thematic saturation before the final participant. The AI provided consistent, literature-grounded scaffolding that supported divergence resolution and accelerated saturation. The HAH-Delphi framework offers a flexible, scalable approach for generating high-quality, context-sensitive consensus. Its successful application across health, coaching, and performance science confirms its methodological robustness and supports its use as a foundation for generating conditional, personalised guidance and published consensus frameworks at scale.

\end{abstract}
\begin{document}

\maketitle
\section{Introduction}

Achieving reliable and nuanced expert consensus is fundamental to advancing evidence-informed practice across the full spectrum of healthcare and applied sciences. From the development of clinical guidelines and rehabilitation protocols to public health recommendations and performance training principles, structured group judgment plays a pivotal role, particularly where empirical evidence is vast, complex, or equivocal, and requires interpretation to become actionable in real-world settings. In such cases, the ability to synthesise informed judgment across experienced professionals is not simply useful, however, it is essential for navigating uncertainty and enabling progress.

The Delphi technique has served as a widely accepted method for structured consensus building for over half a century~\citep{dalkey1963experimental, hasson2000research, linstone1975delphi, niederberger2020delphi, nasa2021delphi, sinha2011using, hall2018recruiting, boulkedid2011using, fink2019delphi, nasa2021delphi}. Originally designed for technological forecasting, it has since become central to health sciences research, underpinning processes from guideline formulation to the identification of research priorities and the definition of clinical competencies. Its defining features, anonymous input to reduce dominance bias, iterative rounds to enable refinement, structured feedback, and statistical aggregation, have ensured its adaptability across diverse contexts. 

However, despite its widespread use and enduring value, the conventional Delphi method is increasingly challenged by a set of practical and methodological limitations that can impair feasibility, diminish insight quality, and reduce the interpretability or applicability of its outputs~\citep{dalkey1963experimental, hasson2000research, linstone1975delphi, niederberger2020delphi, nasa2021delphi, sinha2011using, hall2018recruiting, boulkedid2011using, fink2019delphi, nasa2021delphi, shang2023use}. A persistent barrier lies in the resource demands associated with assembling and managing large expert panels, which in some studies has exceeded 100 participants~\citep{shang2023use}. Sustaining engagement through multiple rounds commonly proves difficult, with attrition rates reported to surpass 90\% in some studies~\citep{shang2023use}. These challenges not only compromise the stability and validity of the resulting consensus but also limit the method’s accessibility for time-constrained or highly specialised experts.

At a methodological level, Delphi processes can inadvertently constrain the very nuance they aim to capture. In pursuit of consensus, responses may converge toward statistical averages or generalised statements, obscuring the conditional logic, domain-specific caveats, and divergent interpretations that define expert-level judgment. Heterogeneous panels may further exacerbate this tendency, amplifying lowest-common-denominator agreement while sidelining minority insights that might otherwise carry significant applied value. Moreover, when consensus is narrowly defined by numerical agreement, it can suppress dissent or misrepresent partial alignment, diminishing the ecological validity of the findings.

The burden placed on facilitators to synthesise qualitative input across large panels and multiple iterations introduces both interpretive load and potential bias. As facilitators necessarily act as intermediaries in shaping feedback summaries, their framing and judgment influence how perspectives are represented back to the panel. Finally, methodological inconsistencies across Delphi studies, particularly around definitions of consensus and treatment of qualitative data, reduce reproducibility and limit comparability between studies~\citep{dalkey1963experimental, hasson2000research, linstone1975delphi, niederberger2020delphi, nasa2021delphi, sinha2011using, hall2018recruiting, boulkedid2011using, fink2019delphi, nasa2021delphi, shang2023use}.

These limitations are not confined to Delphi. Many widely used consensus processes, including modified Delphi formats, structured single-round expert panels, and evidence-based guideline working groups, exhibit similar weaknesses: loss of conditional nuance, inconsistent definitions of consensus, and limited transparency in how expert reasoning is synthesised. Despite procedural differences, these formats often rely on unstructured synthesis and binary agreement metrics, leaving them vulnerable to the same interpretive flattening and lack of reproducibility. A new model must therefore address not only the specific limitations of Delphi, but the broader structural challenges of expert consensus across designs.

Together, these limitations highlight a need for innovation. This study proposes that Generative Artificial Intelligence (GenAI), when used within rigorously defined boundaries and under expert facilitation, offers a powerful means to address these longstanding constraints, not through automation of expert reasoning, but through its strategic augmentation. The HAH-Delphi model developed and evaluated herein introduces a reconfigured approach to consensus generation, one that reconsiders panel structure, redistributes cognitive labour, and systematises the analysis of qualitative insight.

This study introduces a Human–AI Hybrid Delphi (HAH-Delphi) model designed to meet that need. It reconfigures the consensus process to balance AI-supported synthesis with structured expert interpretation, enabling small senior panels to generate rich, conditional, and implementable guidance. The model introduces two core methodological innovations: a structured qualitative saturation framework to assess expert sufficiency, and a four-tier consensus classification system (Strong, Conditional, Operational, Divergent) that moves beyond simplistic percentage thresholds. These components are integrated through structured facilitation, constrained AI synthesis, and justification-based analysis, supporting the generation of context-sensitive, transparent consensus across a range of expert formats.

Advanced GenAI models, such as Gemini 2.5 Pro~\citep{comanici2025gemini}, are uniquely positioned to retrieve, synthesise, and contextualise extensive scientific evidence at a breadth and speed beyond human capability. Their integration into the early stages of the Delphi process relieves senior experts of lower-order retrieval and synthesis tasks, allowing them to concentrate on what AI cannot yet replicate: deep contextual interpretation, application of experiential wisdom, and the articulation of conditional, domain-specific reasoning. This rebalancing supports the use of smaller, more experienced panels without compromising the breadth or depth of the outputs.

A central methodological innovation in this HAH-Delphi model is the use of systematic qualitative analysis to evaluate thematic saturation in expert reasoning. This step, seldom incorporated in traditional Delphi studies, provides a direct mechanism to determine whether the insights generated by a small senior panel are sufficiently comprehensive, offering a grounded measure of expert sufficiency and content completeness. The model further enables structured comparisons across differing levels of expertise, as demonstrated through sub-studies involving less experienced participants, to explore how reasoning depth and alignment vary across groups.

Crucially, the model adheres to the principle of augmentation rather than substitution. All AI-derived material is generated from a constrained corpus of trusted, predominantly public sources, set and  validated by a domain-expert facilitator prior to being made available to the panel. This procedural rigour is designed to mitigate risks such as hallucinated content, overconfidence, or bias propagation, which are issues increasingly recognised in large language models.
This study aimed to develop, implement, and evaluate this HAH-Delphi model, testing its capacity to enhance epistemic clarity, interpretive richness, and operational feasibility within the consensus-building process. In doing so, it explores the redefinition of what constitutes expert input, how panels are composed, and how best to extract meaningful insight from their contributions in an era of human–AI collaboration.

This study aimed to develop, implement, and evaluate the HAH-Delphi framework capable of addressing longstanding limitations in expert consensus generation, particularly those affecting interpretive clarity, conditional nuance, and methodological feasibility across real-world formats. Specifically, we sought to:

\begin{itemize}
\item Validate the synthesis capabilities of generative AI by assessing whether an AI model could replicate the item-level outcomes of six previously published expert consensus sources, including Delphi studies, modified Delphi formats, structured panels, and guideline-derived recommendations, when constrained to using only contemporaneous public evidence.

\item Compare the reasoning depth of human experts and AI by analysing qualitative justifications across a shared set of consensus items within a single domain (chronic insomnia).

\item Test the HAH-Delphi model in applied domains to determine whether small panels of senior experts, supported by AI, could achieve rich, context-sensitive consensus with thematic saturation.

\item Explore the contribution of less experienced experts and assess how their input differed in structure, alignment, and reasoning depth from that of senior experts.

\item The overarching goal was to evaluate whether a compact, structured, and generalisable Human–AI hybrid model could enhance the depth, transparency, and scalability of expert consensus development, across diverse domains and consensus formats.

\end{itemize}
\section{Methods}
\label{sec:methods}

\subsection{Study Design and Phases}

This study employed a three-phase design to develop, test, and evaluate the HAH-Delphi model. Each phase was designed to address a specific methodological aim, progressing from retrospective simulation, through prospective comparison, to applied implementation in complex, real-world domains. We developed and evaluated the HAH-Delphi framework using Gemini 2.5 Pro~\citep{comanici2025gemini}. Throughout this paper, Gemini 2.5 Pro will be abbreviated to Gemini.

\subsection{Phase I (Retrospective Validation)}
\begin{figure*}
    \centering
    \includegraphics[width=0.95\textwidth]{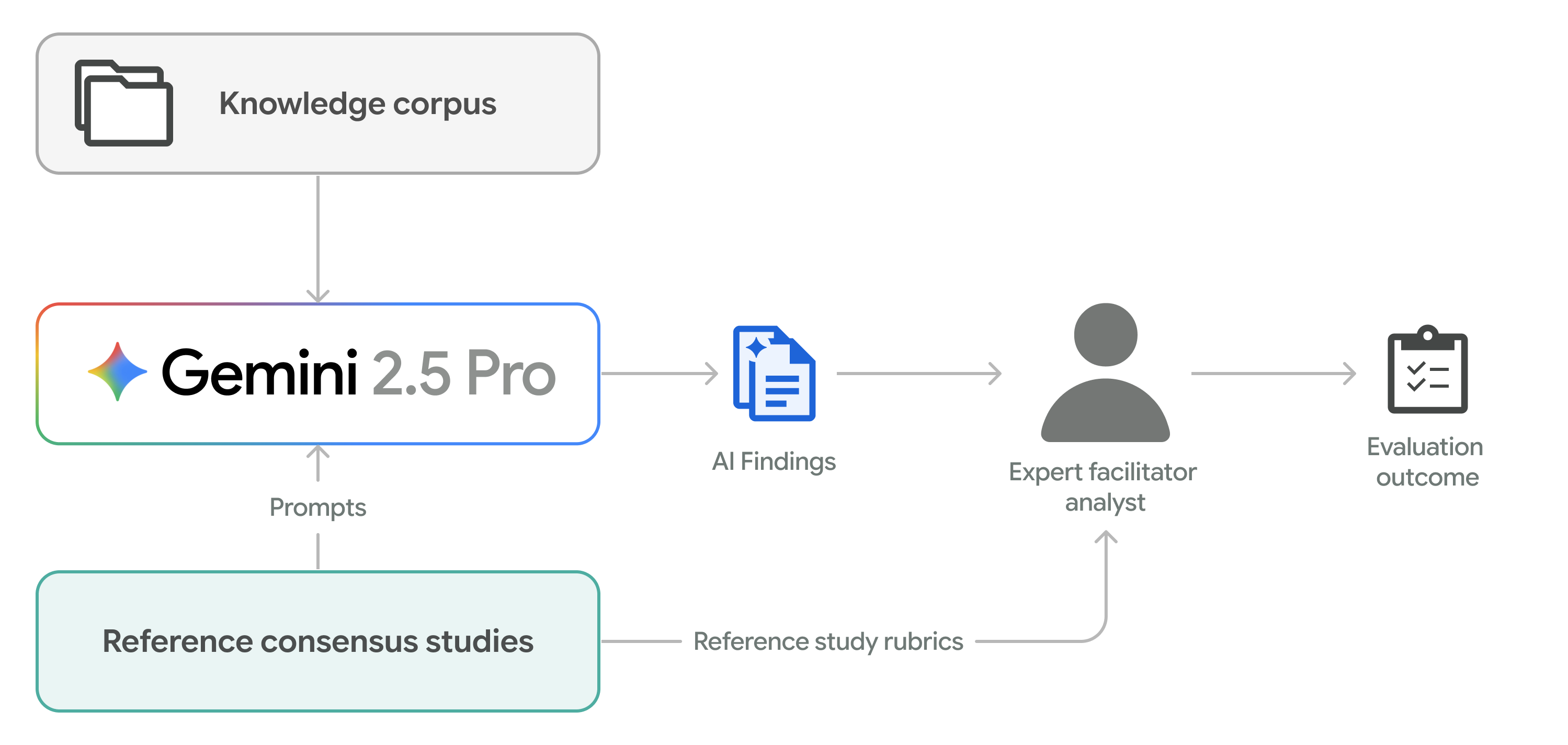}
    \caption{\textbf{Phase I: Retrospective Evaluation.}}
    \label{fig:figure1}
\end{figure*}

Phase I tested whether a generative AI model (Gemini) could replicate item-level outcomes from six previously published expert consensus sources, using only contemporaneous public-domain evidence available at the time of each study (Figure \ref{fig:figure1}).
While formal Delphi studies were a key focus, the included sources also encompassed modified Delphi processes, structured single-round expert panels, and high-grade guideline-derived consensus statements. This methodological diversity was intentional, designed to reflect the varied formats through which expert-derived standards are developed in real-world practice. The six selected studies spanned a range of clinical and applied domains, including insomnia, sedentary behaviour, concussion, low back pain, rotator cuff disorders, and hypertension, and allowed evaluation of the AI’s alignment with both Delphi-specific and broader expert consensus outputs. The corresponding consensus formats were: Delphi (insomnia, rotator cuff), modified Delphi (sedentary behaviour), structured panel (concussion), and guideline-derived consensus (low back pain, hypertension). Full study details, including topic, structure, and consensus type, are provided in Appendix \ref{app:app1} .

In this phase, Gemini was constrained to a predefined corpus of publicly available sources, as specified in Appendix \ref{app:app1}. For panel-based studies, the corpus was restricted to before the first item distribution or expert convening; where this was unknown, a cutoff of at least 9 months prior to publication was applied. For systematic reviews, the cutoff was set to the publication date. It was explicitly prohibited from accessing non-public material, including full-text articles behind paywalls, post-publication summaries, or any content with access restrictions. All AI inputs were structured to mirror the original study formats (e.g. Likert ratings, binary decisions, prioritisation tasks). The expert facilitator ensured corpus fidelity, extracted and reformatted all items, prompted the AI, and analysed outputs for comparison.

\subsection{Phase II (Prospective Study)} 
\begin{figure*}
    \centering
    \includegraphics[width=0.95\textwidth]{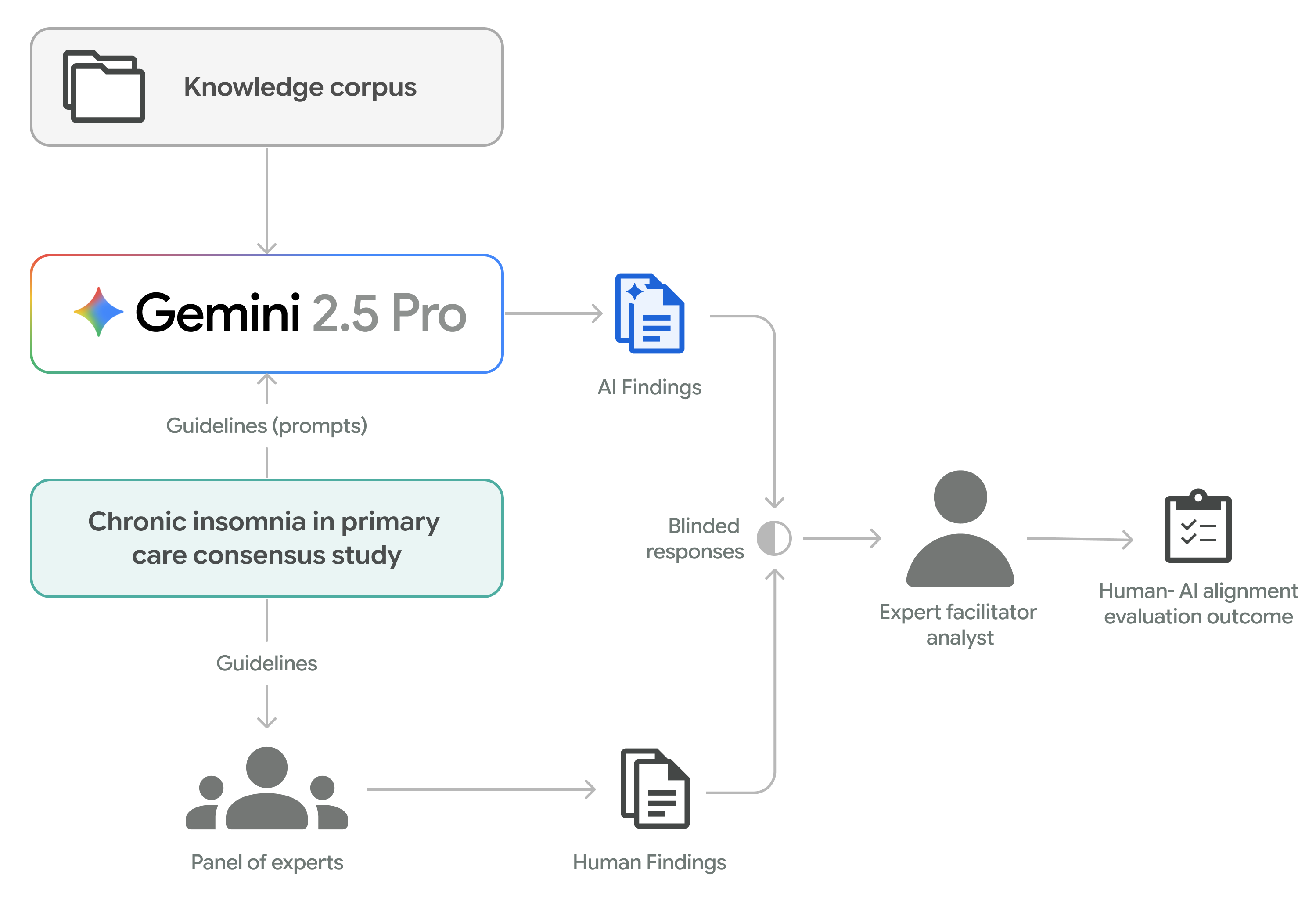}
    \caption{\textbf{Phase II: Prospective Evaluation.}}
    \label{fig:figure2}
\end{figure*}

Phase II re-administered the questions from one of the Phase I benchmark studies, on chronic insomnia in primary care, to a new panel of six human experts (Figure \ref{fig:figure2}). These clinicians were independently contracted as subject matter experts through a third-party medical expert service. Participants rated 20 items using a 5-point Likert scale and provided open-text justifications. Gemini was given the same items and prompted using updated evidence from within its allowed corpus (current to February 2025). All responses were blinded. This phase enabled direct item-level comparison of AI and human responses, including alignment of both conclusions and reasoning structures.

\subsection{Phase III: Applied Delphi in Endurance and Mixed Cardio/Strength Training}
In Phase III, the HAH-Delphi model was implemented in two complex, high-variability training domains: (A) recreational endurance running and (B) resistance and mixed cardio/strength training. Both domains were selected for their relevance to large non-elite populations and their demand for nuanced, context-sensitive application of foundational principles. The primary aim of this phase was to explore how experienced practitioners apply foundational training principles in real-world scenarios, particularly when managing individual variability, life disruptions, and areas of incomplete or ambiguous scientific guidance. A secondary aim of this phase was to explore the methodological implications of incorporating an AI language model as an independent respondent within this HAH-Delphi model. This involved assessing its response patterns against human expert consensus and divergence, and understanding its potential utility in complex knowledge synthesis and consensus-building endeavors. A third aim of this phase was to evaluate the contribution of a panel of less experienced practitioners, whose qualifications would typically meet standard Delphi study inclusion criteria, but who lacked the same depth of applied expertise as the senior panel. Each study was conducted as a structured, single-round Delphi using a common methodological framework and was facilitated by the same expert lead.

\textbf{Definition of Experts}

\textbf{Senior experts} held relevant qualifications and were selected for substantial applied experience, including direct responsibility for program design across varied populations. Selection prioritised real-world decision-making, contextual exposure, and the ability to contribute structured, practice-informed input. Experts were sourced and compensated through a third-party service specializing in providing independent medical expert consultation. \textbf{Less Experienced Participants} are less experienced participants also held relevant qualifications but lacked the applied depth required for senior panel inclusion. They had limited responsibility for independent program design and narrower exposure to diverse coaching or training contexts.

\subsubsection{Phase III.A: Endurance Training Study}
A 143-item questionnaire was developed across 13 thematic domains, covering weekly structure, long run design, intensity zone distribution, tapering, fatigue monitoring, strength integration, rest days, and recovery management. The panel comprised six senior experts (four experienced coaches and two academic practitioners with applied credentials) and eight less experienced participants. All participants rated each item on a 5-point Likert scale and provided detailed justifications. Consensus was assessed using the predefined categories of Strong, Conditional, Operational, and Divergent. Thematic saturation was reached before the sixth expert, and all seven justification types were identified. The structured guidance document containing all consensus principles is included in Appendix \ref{app:app4}.

\subsubsection{Phase III.B: Resistance and Mixed Cardio/Strength Training Study}
This study addressed 159 items across 10 core domains, including progression logic, fatigue management, autoregulation, concurrent training, special populations, volume/intensity regulation, and adaptation over time. The panel again consisted of six senior experts (four coaches, two academic practitioners) and eight less experienced participants. The same consensus classification and justification framework were used. All seven reasoning categories were present across responses, and thematic saturation was reached before expert six. Gemini diverged on four items, each involving contextual judgment that exceeded evidence-based summaries. Consensus outputs from this study are in Appendix \ref{app:app5}.

\subsection{Consensus Categorisation Framework}
Consensus was assessed using a novel integrated framework combining Likert ratings and qualitative reasoning analysis. This allowed classification of agreement not only by score, but by the convergence or divergence of underlying logic.
Conventional Delphi studies typically define consensus using arbitrary percentage thresholds (e.g. $\ge$70–80\% agreement), with limited attention to the reasoning behind those ratings. This approach risks overstating alignment, suppressing valuable dissent, and obscuring conditional or context-specific logic, especially in domains requiring nuanced judgment~\citep{hasson2000research, rahimi2024saturation}. Justifications, if collected, are often analysed narratively and post hoc, rather than used to structure consensus classification itself.
The framework developed in this study addresses these limitations by integrating quantitative agreement with structured qualitative analysis. It introduces four categories, Strong, Conditional, Operational, and Divergent consensus, defined not only by rating convergence but by the coherence and compatibility of expert justifications (Table \ref{tab:table_1}). This allows for clearer differentiation between general agreement and conditional validity, improving interpretability, transparency, and practical relevance. Where clarification revealed that divergent views were in fact conditionally reconcilable, reclassification to 'Conditional Consensus' was permitted. The facilitator adjudicated all such cases.


\begin{table}[h!]
\centering
\caption{Consensus Categories.}
\label{tab:table_1}
\begin{tabularx}{\textwidth}{
    >{\bfseries}l 
    >{\RaggedRight}X   
}
\toprule
Category & Definition \\
\midrule
Strong Consensus &  $\ge$75\% agreement (Likert 4–5 or 1–2), with clearly shared justifications indicating general applicability. \\
\addlinespace 
Conditional Consensus & Varied Likert ratings reconciled by common justifications specifying valid conditions or context-specific boundaries. \\
\addlinespace
Operational Consensus & 67–74\% directional agreement, with the remaining justifications expressing only minor reservations or weak divergence. \\
\addlinespace
No Consensus / Divergent & No coherent agreement on score or logic. Justifications were irreconcilable or conceptually incompatible. \\
\bottomrule
\end{tabularx}
\end{table}

\subsection{Thematic Reasoning Framework}
To assess reasoning structure and depth across respondent types, all free-text justifications from Phases II and III were manually coded by the facilitator using a predefined seven-category thematic framework (Table \ref{tab:table_2}). This framework was developed specifically for the study to capture the layered, conditional, and source-specific nature of expert reasoning in applied domains. It distinguishes between forms of conditionality, general, population-specific, and phased, as well as sources of logic, including evidence-based, experiential, pragmatic, and principle-driven justifications. The framework supports multi-label coding and was applied consistently to both AI and human responses, enabling structured comparison of reasoning breadth, interpretive richness, and justification architecture. Each justification could be assigned more than one category, and in many cases was. Where multiple reasoning types coexisted, a multi-label classification was applied. This approach preserved the depth and structure of expert cognition and allowed the facilitator to identify layered or context-integrated reasoning. The same framework was used to analyse AI and human justifications.


\begin{table}[h!]
\centering
\caption{Thematic Reasoning and Justification Framework}
\label{tab:table_2}
\begin{tabularx}{\textwidth}{
    >{\bfseries}l 
    >{\RaggedRight}X
}
\toprule
Reasoning Type & Definition \\
\midrule
Conditional (General) & Application limited to broadly defined, non-universal conditions. \\
\addlinespace 
Conditional (Population-Based) & Reasoning tied to specific user groups, characteristics, or demographics. \\
\addlinespace
Conditional (Temporal/Phased) & Reasoning tied to time, adaptation stage, or phase-based progression. \\
\addlinespace
Evidence-Based & Justifications citing scientific literature, guidelines, or research synthesis. \\
\addlinespace
Experiential & Based on applied knowledge, observation, or real-world practitioner experience. \\
\addlinespace
Pragmatic & Centred on feasibility, logistical reality, or implementation constraints. \\
\addlinespace
Principle-Based & Founded on theoretical constructs, physiological laws, or foundational training models. \\
\bottomrule
\end{tabularx}
\end{table}

\subsection{Thematic Saturation Assessment}
While thematic saturation is well established in qualitative research~\citep{guest2006many, fusch2015we}, its use in Delphi studies is limited and inconsistently defined~\citep{hasson2000research, junger2017guidance}. Where referenced, it typically reflects item-level redundancy or convergence in scoring, rather than interpretive sufficiency.

This study introduces a more robust and domain-specific application of saturation, grounded in the coverage and convergence of predefined reasoning categories across thematic domains. Unlike generic definitions, this approach captures not only the breadth but also the closure of expert insight, providing a reproducible threshold for panel sufficiency based on reasoning depth rather than response volume.

Thematic saturation was assessed in Phase III to determine whether the compact senior expert panels were sufficient to capture the full spectrum of relevant reasoning across both domains. Crucially, saturation was not defined by the mere absence of new reasoning types, nor by superficial item-level repetition. Instead, it required redundancy across interpretive categories and conditional structures. This represents a more stringent and functionally grounded definition than is typically employed, and is consistent with the model’s emphasis on reasoning depth over numeric sufficiency.

In this study, we defined thematic saturation as the point at which:
\begin{itemize}
    \item All seven predefined reasoning categories (see Table \ref{tab:table_2}) had been represented across the expert justifications;
    \item No new interpretations, qualifiers, or conditional refinements were introduced within those categories that would alter the formulation or applicability of a principle; and
    \item Expert order became immaterial, as reasoning content showed clear convergence and repetition, indicating diminishing returns from additional input.
\end{itemize}

This definition captures not only thematic breadth, but also the stability and closure of expert insight, and was applied consistently to determine panel sufficiency.

The facilitator tracked the presence and distribution of reasoning types across expert responses, mapping them to item counts and distinct thematic domains (e.g., intensity zones, recovery models, progression logic). Saturation required full coverage of reasoning categories across these domains, not simply overall emergence. This approach prevented false positives arising from superficial or isolated mentions.

New experts were considered to contribute meaningfully only if they introduced distinct conceptual approaches, interpretive refinements, or novel contextual reasoning not previously observed. In both applied studies, saturation was achieved before the final expert. The sixth participant added elaborative detail and verbal variation but did not extend the reasoning landscape or shift the interpretive architecture.

Gemini’s responses were similarly reviewed for thematic coverage using this framework, though saturation assessment focused primarily on the human panels. Less-experienced participants were coded using the same system but analysed separately. Their inputs did not expand the reasoning landscape or thematic coverage and were not required to meet saturation thresholds.

\subsection{AI–Human Alignment Classification}
Alignment between Gemini and human expert responses was evaluated independently from saturation. For each item rated by both, the AI’s conclusion and accompanying justification were compared to the consensus and reasoning structure of the expert panel (Table \ref{tab:table_3}). This classification was applied in Phases II and III. The facilitator independently reviewed and adjudicated all alignment calls to ensure rigour.


\begin{table}[h!]
\centering
\caption{AI-Human Panel Alignment Categories}
\label{tab:table_3}
\begin{tabularx}{\textwidth}{
    >{\bfseries}l 
    >{\RaggedRight}X
}
\toprule
Alignment Category & Definition \\
\midrule
Fully Aligned & AI reached the same conclusion and provided a justification closely resembling expert reasoning. \\
\addlinespace 
Partially Aligned & AI endorsed a broadly similar stance but used different or less nuanced logic. \\
\addlinespace
Divergent & AI provided a substantively different conclusion or justification inconsistent with expert consensus. \\
\bottomrule
\end{tabularx}
\end{table}

\subsection{Questionnaire Design and Delphi Execution (Phase III)}
The Delphi questionnaires for the endurance and strength/mixed-modality training studies were developed through structured evidence synthesis by the expert facilitator, supported by Gemini. Items were designed to examine both foundational principles and nuanced areas of disagreement or conditionality, particularly where implementation challenges or contextual variability were anticipated.

In the Endurance Training study, the questionnaire comprised 18 core questions, each accompanied by multiple follow-up questions. These were organised across 13 thematic sections, yielding a total of 179 Likert-scale items. Of these, 143 were fixed statements developed a priori, and 36 were 'Other' items that allowed participants to propose and rate additional alternatives.

In the Strength/Mixed-Modality Training study, the questionnaire was structured into 10 thematic sections and included 159 Likert-scale items, also combining fixed statements and expert-specified alternatives.

All participants, including both senior and less-experienced contributors, completed the full questionnaire and were required to provide written justifications for each rating. The study employed a single-round Delphi format to prioritise depth and quality of reasoning without seeking forced convergence. The expert facilitator reviewed all responses and solicited clarifications where necessary.

\section{Results}
\label{sec:results}

\subsection{Phase I: AI Replication of Historical Consensus}
Gemini replicated human expert consensus with high fidelity across six published expert consensus sources, spanning Delphi studies, modified Delphi formats, structured expert panels, and guideline-derived recommendations~\citep{corp2021evidence, patricios2023consensus, bames2012standardized, requejo2022international, williams20182018, morin2024delphi} (see Appendix \ref{app:app3}). Of 40 core propositions reconstructed from these studies, Gemini reached full alignment with published consensus in 95\% (38/40) and partial alignment in the remainder.  Where there was partial alignment this reflected areas where the literature was actively evolving at the time of the original studies (e.g., shifting pharmacological guidance or emerging diagnostic thresholds). These findings support the feasibility of using a generative AI model to simulate expert consensus retrospectively, based purely on a defined historical evidence base.

This validated the first premise of the HAH-Delphi model: that AI can serve as a transparent, reproducible, and literature-grounded benchmark for consensus development, particularly in domains where synthesising broad evidence is time- or labour-intensive.

Performance was consistent across all study types, including formal Delphi studies, modified Delphi processes, structured expert panels, and high-consensus guideline syntheses, indicating that Gemini was capable of aligning with expert-derived standards regardless of the underlying consensus method. Where divergence occurred, it typically reflected areas of evolving clinical uncertainty or interpretive nuance, rather than limitations specific to any one consensus format. 

\subsection{Phase II: Prospective AI–Human Comparison}
A convened panel of six contracted senior sleep experts completed the 20-item insomnia consensus questionnaire. Gemini was prompted concurrently, using access to publicly available evidence up to December 2023. Likert ratings were analysed for directional concordance, and free-text justifications were evaluated thematically.

\subsubsection{Directional Concordance on Consensus Items}
Gemini demonstrated strong directional concordance with the senior expert panel on the 20-item insomnia questionnaire. Likert-scale ratings were compared at the level of agreement bands (i.e., agree/strongly agree, neutral, disagree/strongly disagree). In 19 out of 20 items (95\%), Gemini's rating fell within the same consensus category as the human panel’s majority stance. This indicated that the AI was able to interpret and apply evidence in a way that closely mirrored the overall direction of expert opinion.

The single item where alignment was not achieved concerned the use of Dual Orexin Receptor Antagonists (DORAs). While human experts provided mixed ratings and contextualised their responses with caution about long-term data and prescribing limitations, Gemini offered an unqualified positive endorsement based on short-term efficacy. This difference in interpretive emphasis led to a classification of divergence, despite some overlapping evidence references.

\subsubsection{Thematic Breadth and Justification Coverage}
All seven predefined reasoning categories were identified in the senior expert justifications. Experts frequently layered multiple reasoning types within single responses, such as combining phased conditionality with experiential reasoning, or embedding pragmatic concerns within evidence-based logic (Table \ref{tab:table_4}). This diversity reflected the contextual complexity of applied decision-making and directly contributed to the richness of the consensus outputs.
In contrast, Gemini consistently applied evidence-based and general conditional reasoning, and occasionally used principle-based logic (Table \ref{tab:table_4}). However, it did not produce experiential or pragmatic justifications, and showed no use of temporal or phased reasoning. These omissions were most evident in items involving implementation trade-offs, behavioural staging, or longitudinal adaptation, contexts that typically require real-world experience or temporally embedded logic

\begin{table}[H]
\centering
\caption{Justification/Reasoning Themes: Senior Experts Vs. Gemini.}
\label{tab:table_4}
\begin{tabular}{ccc}
\hline
\textbf{Reasoning Theme}                & \textbf{Senior Experts} & \textbf{Gemini} \\ \hline
Conditional (General)          & Y              & Y      \\ \hline
Conditional (Population-Based) & Y              & Y      \\ \hline
Conditional (Temporal/Phased)  & Y              & N      \\ \hline
Evidence-Based                 & Y              & Y      \\ \hline
Experiential                   & Y              & N      \\ \hline
Pragmatic                      & Y              & N      \\ \hline
Principle-Based                & Y              & Y      \\ \hline
\end{tabular}
\end{table}

\subsubsection{Thematic Saturation and Panel Sufficiency}
Thematic saturation was assessed to determine whether the six-person senior expert panel was sufficient to capture the full range of relevant reasoning types across the questionnaire. Saturation was defined not simply as the appearance of all seven reasoning categories, but as the point at which additional experts failed to introduce novel interpretive logic, alternative framings, or reasoning structures that would alter the meaning or application of the consensus principles. For saturation to be confirmed, full category coverage had to occur across the questionnaire's thematic domains, and later experts had to offer either elaboration or stylistic variation, but not new conceptual insight.
This threshold was reached by the fifth expert. All seven reasoning types had appeared by expert four, and expert five added unique conditional refinements. Expert six provided additional phrasing and detail but introduced no new reasoning categories or distinct contextual logic. Furthermore, saturation was confirmed as robust to response order: whichever way expert input was sequenced, the cumulative reasoning landscape stabilised prior to the sixth participant.
These findings confirmed that the small, senior panel was sufficient to produce context-rich, interpretively complete guidance under the HAH-Delphi model.

\subsection{Phase III: Applied Evaluation of the HAH-Delphi Model}

\subsubsection{Consensus Achievement and Coverage}
In the Endurance Training study, the panel of six senior experts achieved a high degree of consensus, with 92.3\% (132 out of 143) of the categorised items reaching Strong, Conditional, or Operational Consensus. No Consensus/Divergence occurred on only 7.7\% (11/143) of items. These items primarily related to: (i) how to handle missed sessions, specifically whether they should be skipped, rescheduled, or adapted; (ii) the optimal design of long runs, including whether to specify fixed durations or allow flexible ranges; (iii) the necessity and placement of fixed rest days; and (iv) the use of fatigue monitoring tools and metrics in recreational populations. In each case, the divergence reflected principled differences in emphasis, such as prioritising structure versus flexibility, or simplicity versus precision, rather than fundamental conceptual disagreement.

In the Strength/Mixed-Modality Training study, the panel of six senior experts also achieved a high level of consensus. A substantial majority of the 159 specific topics resulted in Strong Consensus (60 items, 37.7\%), Conditional Consensus (94 items, 59.1\%), or Operational Consensus (5 items, 3.1\%). Items that did not reach consensus were limited in number and typically involved boundary conditions for application, such as the use of training to failure in specific lifts or population-dependent recovery strategies under fatigue. The full consensus principles for resistance and mixed-modality training are likewise published separately as a dedicated guidance document [see Supplementary Resource B].

\subsubsection{Reasoning Structure and Thematic Diversity}
In both studies, all seven predefined reasoning categories, Conditional (General), Conditional (Population-Based), Conditional (Temporal/Phased), Evidence-Based, Experiential, Pragmatic, and Principle-Based, were used by the senior expert panels. These reasoning types were distributed across all thematic sections of each questionnaire. Experts provided detailed, context-aware justifications that incorporated specific variables such as training age, fatigue status, and programme goals. Phrasing frequently included qualifiers such as “when appropriate,” “if fatigue permits,” or “depending on the individual’s context.”
Across both panels, experts drew on practical experience, theoretical principles, and scientific evidence to produce justifications that were both context-sensitive and well-grounded. Most responses integrated multiple reasoning types, reflecting the interpretive complexity of applied programme design. This diversity was central to principle development and was systematically captured by the HAH-Delphi model’s justification analysis.

\subsubsection{Panel Sufficiency and Thematic Saturation}
Thematic saturation was achieved before the sixth expert in both studies. In the Endurance study, all seven predefined reasoning categories had emerged by the fourth expert. The fifth expert added meaningfully distinct variants of conditional logic and application nuance. The sixth expert contributed only redundant or elaborative content. In the Strength/Mixed-Modality study, all reasoning types were covered by expert five, with expert six adding expression and emphasis but no new categories or decision pathways.

Response order had no effect on the emergence of reasoning types. Regardless of the sequence in which experts were analysed, the point at which additional conceptual framings ceased to emerge was consistently reached before the sixth participant.
These results confirm that a small, carefully selected panel of senior experts was sufficient to achieve saturation under the criteria specified in Methods Section IV.

\subsubsection{Role of Gemini in Evidence Benchmarking}
Gemini served as a valuable evidence baseline. Although it lacked the full contextual nuances of senior experts in applied settings, its structured output was useful where human experts diverged. In both studies, Gemini demonstrated a strong capacity to reflect evidence-informed core principles and systematically apply conditional logic, often mirroring the conclusions of academic experts and providing a robust baseline of established knowledge. It contributed to the rapid saturation of foundational knowledge and common conditionalities. Its justifications were typically well-structured and drew from the scientific literature.

However, in the Endurance study, Gemini utilised 'Neutral/Unsure' scores significantly less than typical human expert averages, resulting in more definitive scaled judgments which sometimes diverged from nuanced human consensus summaries, despite its textual justifications often articulating conditional factors. Divergences occurred on topics like the definitive role of wearable technology or specific numerical limits for long runs, where the AI’s stance was more assertive than the more cautious or variable human consensus.

In the Strength/Mixed-Modality study, Gemini diverged from senior expert consensus on 4 out of 159 topics. These included: (1) whether failure-based training should be used for compound lifts in novice populations; (2) the importance of integrating autoregulatory strategies during deload weeks; (3) whether to set fixed repetition ranges for hypertrophy work across diverse populations; and (4) how to prioritise fat loss versus recovery capacity when adjusting training under fatigue. In each case, Gemini’s recommendations were well-structured and grounded in published literature but lacked the applied contextual reasoning provided by senior experts. While not inaccurate in isolation, these recommendations did not reflect the layered decision-making or safety-oriented judgment required for implementation in variable real-world settings.

\subsubsection{Contributions of Less-Experienced Participants}
To assess the potential value and limitations of including lesser-experienced participants in consensus development, their responses were compared to those of the senior expert panels. This analysis focused on both scoring patterns and reasoning quality. 
In the Strength/Mixed-Modality Training study, the larger sample size enabled a structured comparison between the eight lesser-experienced respondents and the senior panel. The lesser-experienced group in this study demonstrated marked differences in both rating behaviour and justification quality. While 82\% of senior expert ratings fell within the 4–5 range and were accompanied by structured, conditional justifications, the less experienced group showed significantly more variability: 38\% of their responses landed at Likert 3, and several respondents selected 1 or 2 on items that were strongly endorsed (Likert 4–5) by the senior panel. Although 65\% of their responses were directionally aligned with the seniors in terms of rating category, only 38\% of their justifications reflected reasoning logic consistent with the expert panel’s conditional framework.

Thematic analysis further confirmed the limits of these contributions. Over half of the less experienced practitioners’ responses contained vague or generic statements, such as “depends on the person” or “progress over time,” without the operational specificity that characterised the senior panel. In areas demanding conditional nuance, such as training to failure, progression strategies, or autoregulation, their justifications were frequently absent, circular, or conceptually inconsistent. Some respondents endorsed a statement with a high Likert score while offering a justification that directly contradicted the intended interpretation or omitted justification entirely.

Several examples highlighted these differences. On progression models, senior experts uniformly referenced autoregulation based on training age; none of the lesser experts mentioned this concept. In concurrent training, all senior experts considered modality type and sequencing to mitigate interference effects, whereas only a quarter of lesser respondents acknowledged modality at all, and none discussed structural programme design. Regarding training to failure, senior experts gave conditional support linked to exercise type, recovery capacity, and training phase, while less experienced respondents often responded with categorical positions lacking context.

In the Endurance Training study, three less-experienced individuals were also included for exploratory comparison. Although no formal statistical analysis was performed due to the smaller sample, the qualitative trends were consistent with those observed in the strength study. Their responses tended to be less precise, less conditional, and notably less detailed than those of the senior panel. No novel reasoning structures were introduced by these participants, and their contributions did not expand the thematic coverage achieved by the senior panel alone. In several cases, their ratings diverged from senior consensus in ways that appeared unanchored to meaningful justification, and their explanatory comments, where present, lacked the interpretive structure required to inform nuanced principle development.

Taken together, these findings demonstrate that divergence between senior and less experienced respondents was evident not only in justification quality but also in scoring reliability. The lack of consistent contextual logic and conditional specificity among lesser participants confirmed that their inclusion risked distorting the clarity and depth of the final outputs. This reinforces a central tenet of the HAH-Delphi model: that the generation of reliable, actionable, and context-sensitive guidance relies on expert panels composed of individuals with the experience and interpretive fluency required to navigate complexity and ambiguity.

\section{Discussion}
\label{sec:discussion}

This study introduced, implemented, and evaluated the HAH-Delphi model designed to overcome the current limitations in expert consensus generation, which include feasibility challenges in recruiting and retaining large panels, the interpretive shallowness of forced agreement models, and the underutilisation of qualitative reasoning in complex domains~\citep{dalkey1963experimental, hasson2000research, linstone1975delphi, niederberger2020delphi, nasa2021delphi, sinha2011using, hall2018recruiting, boulkedid2011using, fink2019delphi, nasa2021delphi, shang2023use}. Across three phases, retrospective replication, prospective comparison, and applied deployment, the HAH-Delphi model demonstrated high methodological transparency, and the capacity to generate rich, context-sensitive consensus outputs from compact panels of senior experts.

Phase I established that a generative AI model (Gemini), constrained solely to pre-publication evidence, could replicate the directional consensus of six historical expert consensus studies, including formal Delphi, modified Delphi, structured consensus, and guideline-derived benchmarks, with 84\% alignment across 32 items. While Delphi studies informed the conceptual design of the HAH-Delphi model, the Phase I benchmarks were intentionally drawn from a broader range of expert consensus approaches. These included not only formal and modified Delphi methods, but also structured single-round expert panels and high-grade guideline-derived recommendations. This diversity reflects the actual landscape of expert knowledge production, particularly in applied health and performance domains, where consensus often emerges through hybrid or pragmatic processes rather than strict methodological templates. That Gemini showed high alignment across this range reinforces the generalisability of the HAH-Delphi framework beyond Delphi-specific replication and supports its application in settings where consensus may take multiple forms. This confirmed the framework’s utility as an evidence synthesiser, capable of inferring the prevailing conclusions of expert panels across multiple consensus formats when appropriately constrained. Importantly, divergence tended to occur in areas where the literature was actively evolving at the time of the original expert consensus, supporting the validity of Gemini’s outputs as reflective of their temporal evidence context.

While Delphi studies formed the methodological foundation of the HAH-Delphi model, Phase I deliberately included a broader range of consensus types. This included structured single-round processes and high-level guideline syntheses where no recent Delphi studies existed. This diversity reflects real-world practice and allows evaluation of AI alignment across differing expert consensus formats. The model thus supports both methodological generalisability and robustness in heterogeneous evidence environments.

Phase II extended this validation to a live setting, comparing Gemini’s responses to those of a newly convened panel of senior sleep experts using a re-administered Delphi questionnaire on insomnia. Gemini aligned with the expert panel’s directional ratings on 95\% of items. However, qualitative analysis revealed key differences: Gemini drew on evidence and structured logic, but it did not replicate the full interpretive richness, experiential insight, or pragmatic reasoning shown by the human experts. Thematic saturation analysis~\cite{malterud2016sample, rahimi2024saturation, junger2017guidance}, not typically performed in Delphi studies,  further demonstrated that all reasoning categories emerged by the fifth expert, reinforcing the sufficiency of a small senior panel. These results highlighted the complementary roles of AI and human experts—AI as a consistent, evidence-aligned foundation, and humans as the source of contextual interpretation and applied judgment.

Phase III applied the full HAH-Delphi model in two demanding domains: endurance training and resistance/mixed-modality training. These were deliberately selected for their complexity, conditionality, and variability. Across both studies, the six-person senior expert panels achieved high consensus (92.3\% in endurance; 159 items categorised in strength/mixed), and thematic saturation was again reached before the sixth expert. Divergence did occur, but it reflected prioritisation trade-offs, e.g., structure vs. adaptability, safety vs. stimulus, not inconsistency or error. These findings directly challenge the assumption that large heterogeneous panels are necessary for credible consensus and demonstrate that compact panels, when carefully selected and facilitated, can generate complete, sophisticated, and implementable guidance.

Notably, the complexity and nuance captured in these studies is not atypical of specialised domains, it is characteristic of them. Most health, rehabilitation, and performance settings require conditional decisions, population-specific logic, and phased application. The failure of many traditional Delphi processes lies not in the effort to find consensus, but in their design’s inability to accommodate nuance without erasing it. By contrast, the HAH-Delphi model elevates nuance as a primary product, not a side-effect, and structures the process accordingly. The successful capture of layered reasoning structures (e.g., temporal logic, general and population-based conditionality, experiential framing) is one of this model’s core strengths.

A second methodological innovation of this study lies in the development and application of a structured, four-tier consensus classification framework, comprising Strong, Conditional, Operational, and Divergent consensus. Unlike conventional Delphi approaches, which rely primarily on percentage agreement thresholds, this framework integrates quantitative convergence with qualitative justification analysis. This dual-anchored model allows for fine-grained distinctions between general agreement, conditional validity, operational sufficiency, and irreconcilable disagreement. It avoids false consensus, preserves conditional nuance, and provides a clear basis for facilitator adjudication. Crucially, it enables structured, implementable principles to emerge even in areas of partial alignment, offering a more transparent and practice-ready alternative to binary consensus/no-consensus models.

Thematic saturation was assessed using a predefined framework covering seven reasoning types. The model required not just category coverage but interpretive redundancy across thematic domains. By the fifth expert, no new reasoning types emerged that would alter principle construction or shift their application boundaries. The sixth expert's input added elaboration but no new conceptual logic. This consistency, verified both in Phase II and III, supports a principled, rather than numerical, definition of panel sufficiency.

The key findings of this work is that Gemini consistently aligned with human experts on core content and exhibited stable, coherent reasoning grounded in the literature. That it did not offer experiential or pragmatic justifications is neither surprising nor problematic at this stage of model development. These reasoning types require situational pattern recognition, value trade-offs, ethical caution, and accumulated applied wisdom, capacities not encoded in literature alone. However, with increased access to structured, high-quality, expert-derived outputs such as those generated in this study, future AI models will likely become more adept at approximating this style of conditional logic. In this way, the HAH-Delphi model not only uses AI, it contributes to its training and evolution.

Importantly, even where Gemini diverged from human experts, its responses remained internally coherent, logically sound, and free of conceptual or safety-critical flaws. Divergence most often reflected simplification, lack of contextual sensitivity, or overconfidence, factors that make it a valuable benchmark but not a replacement for human oversight. The AI’s role in areas of human disagreement was often stabilising: by offering a neutral, evidence-based scaffold, it helped crystallise key distinctions and prompted clearer articulation of expert reasoning.

The role of the expert human facilitator was central throughout. This individual was not a neutral intermediary but a highly skilled interpreter responsible for ensuring internal coherence, resolving ambiguities, aligning Likert ratings with nuanced justifications, and conducting thematic synthesis. Facilitator-led clarification was especially important where text justifications were conditionally framed or misaligned with scalar ratings. The rigour and clarity of the final principles were directly dependent on this role, which should be considered a requirement for successful implementation of the HAH-Delphi model.

The inclusion of less-experienced respondents provided important contrast. Their responses, though generally aligned, lacked the structured conditionality, applied nuance, and interpretive depth seen in the senior panels. They did not introduce new reasoning categories and, if weighted equally, would have diluted the thematic precision of the final outputs. This does not suggest that less experienced perspectives are without value, but rather that when the objective is to generate deeply contextualised and conditional guidance, epistemic quality must take precedence over numerical inclusiveness.

The one-round structure of Phases II and III placed high initial demands on experts, but it avoided the attrition and procedural fatigue associated with multi-round designs. Structured facilitation, expert familiarity with the domain, and the inclusion of Gemini as an evidence scaffold allowed the full range of reasoning to be elicited in a single round. This format proved not only feasible but efficient, and particularly suitable for settings in which expert availability is limited or where rapid guidance development is necessary.

In summary, the HAH-Delphi model offers a viable, flexible, and rigorous approach to expert consensus development in complex domains. It does not seek to automate expert judgment but to structure, preserve, and elevate it, integrating the strengths of generative AI, senior expert reasoning, and skilled human facilitation. The findings from all three phases support its application across a range of domains where nuanced, conditional, and practically applicable guidance is required. As AI systems continue to evolve and consensus-building becomes more central to health and performance decision-making, the importance of such a structured hybrid model is likely to grow.

\subsection{Limitations}
While this study provides strong support for the HAH-Delphi model, limitations and operational considerations should be acknowledged.

First, although the AI model (Gemini) performed reliably as an evidence synthesiser and benchmark across all phases, its outputs remain constrained by current model architecture and its dependence on textual sources. It did not produce experiential, pragmatic, or ethically grounded justifications, reasoning types that are central to expert decision-making. This limitation is expected, given the nature of generative models, and it does not undermine the AI’s role as a structured, transparent scaffold. However, it reinforces the model’s dependence on human interpretive input and facilitator oversight.

Second, the model’s success depends heavily on expert selection. While this was operationalised through strict inclusion criteria, the ability to identify and recruit truly reflective senior experts remains a key dependency. Additionally, the human experts participating in Phases Two and Three were compensated for their time and expertise which introduces a potential source of bias. This is not a limitation of the model per se but an implementation requirement: if high-quality, contextualised outputs are the goal, panel selection must prioritise epistemic depth over surface-level representativeness. Broader stakeholder engagement may be appropriate in other consensus contexts, but not when generating nuanced guidance. Moreover, the number of expert raters in this study was relatively small, so the model's generalization and reliability with a larger panel require validation in future studies.

Third, the role of the facilitator is crucial and requires significant depth and breadth of expertise. Thematic synthesis and clarification processes demand consistency, interpretive judgment, and methodological discipline. While the use of a single senior facilitator may be seen as a limitation, it reflects standard practice in consensus research, especially Delphi format,  where one lead is typically responsible for guiding item development, coordinating panel interaction, and adjudicating responses. The model adopted here was consistent with published guidance and methodological norms. In real-world settings, the feasibility of deploying multiple facilitators with equivalent expertise is low, particularly in specialist domains requiring subject-matter fluency, methodological rigour, and continuity of interpretation. Importantly, this study mitigated bias risks by using pre-specified reasoning categories, transparent adjudication processes, and explicit documentation of facilitator decisions across all phases. Nonetheless, future studies should test the model with several different facilitators to validate its facilitator-independent reliability.

Finally, the one-round structure, though effective in this study, may not be suitable in all cases. It relies on robust item construction, responsive clarification procedures, and cognitively prepared panels. In domains where item clarity is low or expert familiarity is limited, a second refinement round may improve outcome quality. Future work should explore when and how to best deploy single- versus multi-round formats under the HAH-Delphi model.

\subsection{Future Directions}

The successful development and application of the HAH-Delphi model opens several avenues for future research, methodological refinement, and domain-specific implementation.

First, further application across diverse domains is essential. While this study focused on applied health and performance contexts, where conditionality and pragmatic nuance are core features, the model is equally relevant to clinical guideline development, rehabilitation planning, public health policy, digital therapeutics, and regulatory consensus. Future research should explore how the model performs in ethically charged, interdisciplinary, or contested domains where evidence alone is insufficient to guide practice, and where reasoning quality becomes the decisive factor in shaping consensus.

Second, additional investigation is warranted into the optimal configuration of AI constraints and interaction design. Although this study used a conservative, facilitator-vetted constraint corpus and static AI benchmarking, future iterations might explore dynamic AI-human interaction loops, version comparisons, or prompt variations to test how generative models can adaptively support, not just benchmark, expert deliberation. This includes examining how future AI models, trained on structured human justifications like those generated here, might evolve to offer more conditional and contextually embedded reasoning.

Third, longer-term follow-up of consensus outputs generated using the HAH-Delphi model could offer insights into their practical impact. Are they more actionable? Do they reduce implementation ambiguity? Are they better received by clinicians, practitioners, or decision-makers than traditional consensus statements? Structured evaluations of downstream uptake and effect would help establish the model not just as a methodological innovation but as a meaningful contributor to practice change.

Fourth, future work should continue to refine definitions of thematic saturation and panel sufficiency in hybrid models. As new AI capabilities emerge, and as expert–AI collaboration becomes more routine, it will be important to revisit and adapt these thresholds to maintain standards of clarity, richness, and sufficiency in expert consensus development.

Finally, a particularly promising direction lies in the deliberate use of structured, high-quality outputs from senior experts to inform the future of AI training. As generative models evolve, their capacity to emulate context-sensitive reasoning will depend heavily on exposure to well-articulated, experientially grounded, and conditionally nuanced expert logic, precisely the type of content produced through the HAH-Delphi process. Future research could explore how curated datasets from hybrid Delphi studies might contribute to the development of more trustworthy, responsible, and domain-specific AI systems capable of supporting complex reasoning under uncertainty. In this respect, senior human experts are not just irreplaceable within the consensus process, they are essential architects of the next generation of AI reasoning.

\section{Conclusion}
This study introduced and evaluated the HAH-Delphi model capable of generating nuanced, actionable, and context-sensitive expert consensus. Across three distinct phases, retrospective validation, prospective comparison, and applied deployment, the HAH-Delphi model consistently demonstrated high alignment between constrained generative AI outputs and human expert reasoning, while preserving the interpretive richness and conditional depth typically absent from traditional Delphi methods.

Small panels of carefully selected senior experts, supported by an evidence-synthesising AI and guided by a domain-competent facilitator, proved sufficient to achieve thematic saturation and principled consensus across complex domains. The model’s structured integration of AI and human input allowed it to overcome common Delphi limitations: high panel burden, attrition risk, methodological opacity, and loss of nuance. At the same time, it preserved and amplified the contributions that only human experts can provide, experiential reasoning, ethical caution, and applied judgment.

While the AI model could not independently replicate the full reasoning spectrum of human experts, it provided a stable, transparent scaffold and helped accelerate knowledge synthesis. Its structured divergence, when present, was informative rather than erroneous, reinforcing the model’s capacity to guide, not replace, expert deliberation. The role of the human facilitator proved central to ensuring interpretive fidelity, methodological consistency, and thematic clarity.

The HAH-Delphi model offers a transparent, flexible, and rigorous pathway to expert consensus formation in domains where guidance must reflect complexity, context, and conditionality. It stands not as a replacement for expert thinking, but as a framework to preserve and amplify it, with clarity, depth, and integrity. Crucially, by systematically capturing the reasoning of senior human experts in structured form, the model creates a valuable corpus for future AI development. As AI systems advance, they will require not just data, but principled, context-rich reasoning to emulate. In this regard, expert panels operating within models such as HAH-Delphi are not only consensus-builders, but teachers—imparting the interpretive scaffolds from which future decision-support systems can learn. As both the challenges and capabilities of human–AI collaboration evolve, models like this will be essential to ensuring that future tools remain both evidence-informed and experience-grounded.

\section{Acknowledgement}

We are deeply grateful to the members of the Consumer Health Research Team at Google for their valuable feedback and technical support throughout this study, in particular Daniel Roggen, Jacqueline Shreibati, Conor Heneghan, Yun Liu, Nova Hammerquist, Brent Winslow, Heiko Maiwand, Mark Malhotra, Shwetak Patel, Mark Brooke, Yojan Patel, Robert Harle, and Florence Thng.



\section{Competing Interests}
This study was funded by Google LLC. C.S., A.A.M. are employees of Alphabet and may own stock as part of the standard compensation package.



\bibliography{delphi}
\pagebreak


\begin{center}
    \huge\bfseries APPENDICES
\end{center}






\appendix

\section{Benchmark Study Description}
\label{app:app1}

\begin{table}[H]
\centering

\begin{tabularx}{\textwidth}{
    >{\raggedright}p{2.6cm} 
    >{\raggedright}p{2.2cm} 
    >{\RaggedRight}X         
    >{\RaggedRight}X         
    c                        
    >{\RaggedRight}X         
}
\toprule
\textbf{Consensus Type} & \textbf{Topic} & \textbf{Study Name} & \textbf{Experts} & \textbf{Rounds} & \textbf{Notes} \\
\midrule

\multirow{2}{=}{\textbf{Delphi}}
& *Insomnia & \citep{morin2024delphi, Lessard2024Insomnia} & 16 multidisciplinary sleep medicine experts. National survey items extracted from Delphi & 2 & Consensus on 37 recommendations using $\ge$75\% agreement threshold. \\ 
\cmidrule(l){2-6} 
& Rotator cuff & \citep{requejo2022international} & 15 & 3 & Formal Delphi process. \\ 
\midrule

\textbf{Modified Delphi} & Sedentarism & \citep{tremblay2017sedentary} & 53 Delphi participants & 2 & Terminology defined using structured feedback rounds. \\ 
\midrule

\textbf{Structured Single-Round Consensus} & Concussion & \citep{patricios2023consensus} & 29 expert authors; 600+ conference attendees reviewed and voted & 1 (conference vote) & Final statements based on $\ge$80\% agreement at 6th Concussion in Sport Conference. \\ 
\midrule

\multirow{2}{=}{\textbf{Guideline-Derived Consensus}}
& Low Back Pain (Primary Care) & \cite{corp2021evidence} & Guideline working group authorship (not applicable) & N/A & Synthesises expert consensus across European guidelines; no fixed panel or Delphi process. \\ 
\cmidrule(l){2-6}
& Hypertension & \citep{williams20182018} & Multidisciplinary panels across documents (not reported) & N/A & Structured expert panels using GRADE; no Delphi methodology used. \\
\bottomrule
\end{tabularx}
\end{table}

*One of the benchmark domains—chronic insomnia—used question items adapted from~\citep{Lessard2024Insomnia}, a national survey developed to assess real-world alignment with expert recommendations from a four-round Delphi consensus~\citep{morin2024delphi}. Both studies were conducted by the same research group, with the~\citep{Lessard2024Insomnia} survey directly reflecting the questions posed and consensus statements generated in the~\citep{morin2024delphi} Delphi. The national findings broadly supported the expert consensus, confirming key priorities while highlighting gaps in adoption. Using these items enabled benchmarking against Delphi-derived content in a format consistent with clinical interpretation and practice-facing decision-making. 
\newpage

\section{Source Corpus \& Boundaries Framework Template for Gemini 2.5 Pro Information Synthesis and Benchmarking.}

\label{app:app2}


\begin{table}[h!]
\centering
\begin{tabularx}{\textwidth}{
    >{\bfseries}p{3.5cm} 
    >{\RaggedRight}X       
}
\toprule
Framework Section & Description / Components (Emphasis on Publicly Available \& Open Access) \\
\midrule

1. Study Topic \& Scope & Defines the specific focus: Topic area, Target Population, Key Questions/Domains covered, and the relevant Knowledge Cutoff Date (for historical simulations or defining currency for live studies). \\
\addlinespace

2. Source Categories \& Inclusion Criteria & 
Specifies allowed information sources, prioritizing publicly available and openly accessible content:
\begin{itemize}[leftmargin=*, nosep]
    \item \textbf{Open Access Peer-Reviewed Literature:} Articles from open access journals or repositories (e.g., PubMed Central, arXiv), noting pre-print status where applicable.
    \item \textbf{Publicly Available Clinical Practice Guidelines (CPGs):} CPGs from major societies (e.g., WHO, NICE, ACSM) available without subscription.
    \item \textbf{Governmental \& Public Health Agency Reports:} Publicly available reports from bodies like the CDC, NHS, etc.
    \item \textbf{Facilitator-vetted Reputable Websites:} Information from pre-approved, evidence-based sites (e.g., professional organizations, academic institutions).
    \item \textbf{Explicit Exclusions:} Paywalled articles, commercial textbooks, social media, forums, and personal blogs.
\end{itemize} \\
\addlinespace

3. Trustworthiness Hierarchy \& Weighting & 
Establishes levels of evidence among publicly available sources:
\begin{itemize}[leftmargin=*, nosep]
    \item \textbf{Level 1:} Major public guidelines, high-quality open access Systematic Reviews/Meta-Analyses.
    \item \textbf{Level 2:} Open access RCTs, public consensus statements.
    \item \textbf{Level 3:} Open access observational studies, vetted professional websites.
    \item \textbf{Level 4:} Other vetted grey literature.
    \item \textbf{AI Instruction:} Prioritize higher levels and note evidence strength.
\end{itemize} \\
\addlinespace

4. Search Strategy Guidance & Provides examples of keywords, synonyms, MeSH terms, and database filters (e.g., "open access," publication type, dates) to guide AI retrieval from the public corpus. \\
\addlinespace

5. Quality Assessment Protocol (Public/Grey Lit Vetting) & Outlines facilitator criteria for vetting grey literature: Authority, Transparency, Objectivity, Referencing, and Currency. \\
\addlinespace

6. Synthesis Rules \& Instructions for AI & Defines AI directives: Adhere to scope; Synthesize objectively from public sources only; Report conflicts; Do not extrapolate; Apply safety filters; Justify conclusions; Paraphrase and attribute to respect IP. \\
\addlinespace

7. Facilitator Validation Criteria & Lists criteria for checking the AI's output: Relevance, Accuracy, Completeness, Nuance, Source Adherence, IP Compliance, and Safety/Bias check. Iteration occurs if checks fail. \\
\bottomrule
\end{tabularx}
\end{table}

\newpage

\section{Summary of Phase I Benchmark Studies and Consensus Methods.}
\label{app:app3}

\begin{table}[H]
\centering
\begin{tabularx}{\textwidth}{
    l
    >{\RaggedRight}X 
    >{\RaggedRight}X
}
\toprule
\textbf{Reference} & \textbf{Topic} & \textbf{Classification in Paper} \\
\midrule
\textbf{\citep{bames2012standardized}} & Sedentary behaviour guidelines & Modified Delphi \\
\addlinespace 
\textbf{\citep{requejo2022international}} & Rotator Cuff signs & Delphi study \\
\addlinespace
\textbf{\citep{patricios2023consensus}} & Concussion return-to-play protocol & Structured panel \\
\addlinespace
\textbf{\citep{corp2021evidence}} & Low back pain recommendations & Guideline-derived consensus \\ 
\addlinespace
\textbf{\citep{williams20182018}} & Hypertension management & Guideline-derived / hybrid consensus \\
\addlinespace
\textbf{\citep{morin2024delphi, Lessard2024Insomnia}} & Insomnia management & Delphi study \\
\bottomrule
\end{tabularx}
\end{table}

\subsection{Chronic Insomnia in Primary Care – Published responses vs. Gemini \citep{morin2024delphi}}

\begin{table}[H]
\centering
\label{tab:appendix3A}
{
\small 
\renewcommand{\arraystretch}{0.9} 

\begin{tabularx}{\textwidth}{
    c              
    >{\RaggedRight}X  
    >{\RaggedRight}X  
    >{\RaggedRight}X  
    c              
    l              
}
\toprule
\textbf{Num.} & \textbf{Question Summary} & \textbf{Gemini Likert Rating} & \textbf{Human Consensus Category} & \thead{\textbf{Human}\\\textbf{Agreement}\\\textbf{Range}} & \textbf{Alignment} \\
\midrule
1 & Take structured sleep history & 5 – Strongly Agree & Strong Agreement & $\ge$90\% 4–5 & Aligned \\
2 & Use validated questionnaires (e.g. ISI) & 5 – Strongly Agree & Strong Agreement & $\ge$90\% 4–5 & Aligned \\
3 & Screen high-risk patients & 4 – Agree & Moderate Agreement & 75–89\% 4–5 & Aligned \\
4 & Use sleep diaries & 5 – Strongly Agree & Strong Agreement & $\ge$90\% 4–5 & Aligned \\
5 & Consider insomnia in comorbid presentations & 5 – Strongly Agree & Strong Agreement & $\ge$90\% 4–5 & Aligned \\
6 & PCPs can deliver brief CBT-I & 4 – Agree & Moderate Agreement & 75–89\% 4–5 & Aligned \\
7 & Refer for full CBT-I if possible & 5 – Strongly Agree & Strong Agreement & $\ge$90\% 4–5 & Aligned \\
8 & Avoid pharmacotherapy as first-line & 4 – Agree & Moderate Agreement & 75–89\% 4–5 & Partially Aligned \\
9 & CBT-I should be first-line & 5 – Strongly Agree & Strong Agreement & $\ge$90\% 4–5 & Aligned \\
10 & DORAs appropriate in select cases & 3 – Neutral & Mixed Consensus & $<$75\% 4–5 & Aligned \\
11 & Sleep hygiene alone is insufficient & 5 – Strongly Agree & Strong Agreement & $\ge$90\% 4–5 & Aligned \\
12 & Use shared decision-making & 5 – Strongly Agree & Strong Agreement & $\ge$90\% 4–5 & Aligned \\
13 & Follow-up is necessary & 5 – Strongly Agree & Strong Agreement & $\ge$90\% 4–5 & Aligned \\
14 & Educate on sleep–wake regulation & 5 – Strongly Agree & Strong Agreement & $\ge$90\% 4–5 & Aligned \\
15 & Advise on caffeine, alcohol, screens & 5 – Strongly Agree & Strong Agreement & $\ge$90\% 4–5 & Aligned \\
16 & Deliver brief advice in routine visits & 4 – Agree & Moderate Agreement & 75–89\% 4–5 & Aligned \\
17 & Use sleep restriction with caution & 4 – Agree & Moderate Agreement & 75–89\% 4–5 & Aligned \\
18 & Refer complex cases to specialist & 5 – Strongly Agree & Strong Agreement & $\ge$90\% 4–5 & Aligned \\
19 & Use digital CBT-I apps & 4 – Agree & Moderate Agreement & 75–89\% 4–5 & Aligned \\
20 & Insomnia should not be dismissed & 5 – Strongly Agree & Strong Agreement & $\ge$90\% 4–5 & Aligned \\
\bottomrule
\end{tabularx}
} 
\end{table}

\subsection{Low Back Pain Management (Primary Care) Source: ~\citep{corp2021evidence}}

\begin{table}[H]
\centering
\label{tab:appendix3B}
\begin{tabularx}{\textwidth}{
    c              
    >{\RaggedRight}X  
    >{\RaggedRight}X  
    >{\RaggedRight}X  
    c              
    l              
}
\toprule
\textbf{Num} & \textbf{Question Summary} & \textbf{Gemini Likert Rating} & \textbf{Human Consensus Category} & \thead{\textbf{Human}\\\textbf{Agreement}\\\textbf{Range}} & \textbf{Alignment} \\
\midrule
1 & Reassure patients; LBP is common and not serious & 5 – Strongly Agree & Strong Agreement & $\ge$90\% 4–5 & Aligned \\
\addlinespace 
2 & Imaging is rarely required & 5 – Strongly Agree & Strong Agreement & $\ge$90\% 4–5 & Aligned \\
\addlinespace
3 & Use NSAIDs first-line; not paracetamol & 4 – Agree & Moderate Agreement & 75–89\% 4–5 & Partially Aligned* \\
\addlinespace
4 & Recommend physical activity and self-management & 5 – Strongly Agree & Strong Agreement & $\ge$90\% 4–5 & Aligned \\
\addlinespace
5 & Avoid bed rest & 5 – Strongly Agree & Strong Agreement & $\ge$90\% 4–5 & Aligned \\
\addlinespace
6 & Avoid routine opioids & 5 – Strongly Agree & Strong Agreement & $\ge$90\% 4–5 & Aligned \\
\bottomrule
\end{tabularx}
\end{table}

Note on Item 3: Gemini, synthesizing pre-2021 evidence, reflected evolving guideline recommendations that increasingly questioned paracetamol’s effectiveness. The Delphi panel showed moderate agreement—likely due to real-world variability and slower de-implementation in practice. Thus, the direction was aligned, but confidence levels differed.

\subsection{Concussion in Sport – Human Experts vs. Gemini
Source: \citep{patricios2023consensus} – Pre-Oct 2022 corpus cutoff applied for Gemini synthesis.}

\begin{table}[H]
\centering
\label{tab:appendix3C}
\begin{tabularx}{\textwidth}{
    c              
    >{\RaggedRight}X  
    >{\RaggedRight}X  
    >{\RaggedRight}X  
    c              
    l              
}
\toprule
\textbf{Num.} & \textbf{Question Summary} & \textbf{Gemini Likert Rating} & \textbf{Human Consensus Category} & \thead{\textbf{Human}\\\textbf{Agreement}\\\textbf{Range}} & \textbf{Alignment} \\
\midrule
1 & Remove athlete from play immediately upon suspicion & 5 – Strongly Agree & Strong Agreement & $\ge$90\% 4–5 & Aligned \\
\addlinespace 
2 & Use a multidimensional assessment approach & 5 – Strongly Agree & Strong Agreement & $\ge$90\% 4–5 & Aligned \\
\addlinespace
3 & Encourage gradual return-to-learn and return-to-sport & 5 – Strongly Agree & Strong Agreement & $\ge$90\% 4–5 & Aligned \\
\addlinespace
4 & Initiate symptom-limited activity early in recovery & 4 – Agree & Strong Agreement & $\ge$90\% 4–5 & Aligned \\
\addlinespace
5 & Use SCAT-type tools to assist with sideline/clinic assessment & 4 – Agree & Strong Agreement & $\ge$90\% 4–5 & Partially Aligned \\
\bottomrule
\end{tabularx}
\end{table}

Note on Item 5: Gemini agreed with using structured tools like SCAT but did not specifically name SCAT6, which was finalized shortly after the cutoff. The alignment is considered valid, as SCAT was already an established tool; the specific version evolution (SCAT6) fell outside the AI’s reference window.

\subsection{Rotator Cuff Signs \citep{requejo2022international}}

\begin{table}[H]
\centering
\label{tab:appendix3D}
\begin{tabularx}{\textwidth}{
    c              
    >{\RaggedRight}X  
    >{\RaggedRight}X  
    >{\RaggedRight}X  
    c              
    l              
}
\toprule
\textbf{Num.} & \textbf{Question Summary} & \textbf{Gemini Likert Rating} & \textbf{Human Consensus Category} & \thead{\textbf{Human}\\\textbf{Agreement}\\\textbf{Range}} & \textbf{Alignment} \\
\midrule
1 & Pain with overhead/abduction movements & 5 – Strongly Agree & Strong Agreement & $\ge$90\% 4–5 & Aligned \\
\addlinespace
2 & Pain localised to deltoid region & 4 – Agree & Strong Agreement & $\ge$90\% 4–5 & Aligned \\
\addlinespace
3 & Age over 40 typical feature & 4 – Agree & Strong Agreement & $\ge$90\% 4–5 & Aligned \\
\addlinespace
4 & Symptom onset after increased activity & 5 – Strongly Agree & Strong Agreement & $\ge$90\% 4–5 & Aligned \\
\addlinespace
5 & Imaging only if trauma/malignancy & 4 – Agree & Strong Agreement & $\ge$90\% 4–5 & Aligned \\
\addlinespace
6 & Imaging not sole basis for surgery decision & 4 – Agree & Strong Agreement & $\ge$90\% 4–5 & Aligned \\
\addlinespace
7 & Assess active shoulder ROM & 5 – Strongly Agree & Strong Agreement & $\ge$90\% 4–5 & Aligned \\
\addlinespace
8 & Assess shoulder strength & 5 – Strongly Agree & Strong Agreement & $\ge$90\% 4–5 & Aligned \\
\addlinespace
9 & Pain on resisted abduction & 5 – Strongly Agree & Strong Agreement & $\ge$90\% 4–5 & Aligned \\
\addlinespace
10 & Pain on resisted external rotation & 5 – Strongly Agree & Strong Agreement & $\ge$90\% 4–5 & Aligned \\
\addlinespace
11 & Special tests unnecessary for diagnosis & 2 – Disagree & Strong Agreement & $\ge$90\% 4–5 & Divergent \\
\addlinespace
12 & Reproducible loading pain sufficient for diagnosis & 3 – Neutral & Strong Agreement & $\ge$90\% 4–5 & Partial \\
\bottomrule
\end{tabularx}
\end{table}

\subsection{Hypertension Diagnosis and Management Pathway – Human Experts vs. Gemini \citep{williams20182018} – Pre 2018 corpus cutoff applied for Gemini synthesis.}

\begin{table}[H]
\centering
\label{tab:appendix3E}
\begin{tabularx}{\textwidth}{
    c              
    >{\RaggedRight}X  
    >{\RaggedRight}X  
    >{\RaggedRight}X  
    c              
    l              
}
\toprule
\textbf{Num} & \textbf{Question Summary} & \textbf{Gemini Likert Rating} & \textbf{Human Consensus Category} & \thead{\textbf{Human}\\\textbf{Agreement}\\\textbf{Range}} & \textbf{Alignment} \\
\midrule
1 & Use ABPM or HBPM to confirm hypertension diagnosis & 5 – Strongly Agree & Strong Agreement & $\ge$90\% 4–5 & Aligned \\
\addlinespace 
2 & Recommend lifestyle changes to all patients & 5 – Strongly Agree & Strong Agreement & $\ge$90\% 4–5 & Aligned \\
\addlinespace
3 & Follow stepwise medication algorithm & 5 – Strongly Agree & Strong Agreement & $\ge$90\% 4–5 & Aligned \\
\addlinespace
4 & Consider lower BP targets in high-risk patients & 4 – Agree & Moderate Agreement & 75–89\% 4–5 & Partially Aligned \\
\addlinespace
5 & Repeat BP measurements and review response & 5 – Strongly Agree & Strong Agreement & $\ge$90\% 4–5 & Aligned \\
\bottomrule
\end{tabularx}
\end{table}

Note on Item 4: Gemini rated lower targets as appropriate but with more caution than the panel. While the Delphi group leaned toward lower BP targets in high-risk individuals, this was a debated issue. Gemini appropriately reflected this ambiguity in the literature of that time.

\subsection{Sedentary Behaviour Terminology – Human Experts vs. Gemini
Source: \cite{tremblay2017sedentary} – Pre2017 corpus cutoff applied for Gemini synthesis.}

\begin{table}[H]
\centering
\caption{Alignment on Sedentary Behaviour Definitions}
\label{tab:appendix3F}
\begin{tabularx}{\textwidth}{
    c              
    >{\RaggedRight}X  
    >{\RaggedRight}X  
    >{\RaggedRight}X  
    c              
    l              
}
\toprule
\textbf{Num.} & \textbf{Question Summary} & \textbf{Gemini Likert Rating} & \textbf{Human Consensus Category} & \thead{\textbf{Human}\\\textbf{Agreement}\\\textbf{Range}} & \textbf{Alignment} \\
\midrule
1 & Sedentary behaviour should be defined by energy expenditure ($\le$1.5 METs) & 5 – Strongly Agree & Strong Agreement & $\ge$90\% 4–5 & Aligned \\
\addlinespace 
2 & Sedentary behaviour should include sitting/reclining posture & 5 – Strongly Agree & Strong Agreement & $\ge$90\% 4–5 & Aligned \\
\addlinespace
3 & The term “physical inactivity” should be distinct from “sedentary behaviour” & 5 – Strongly Agree & Strong Agreement & $\ge$90\% 4–5 & Aligned \\
\addlinespace
4 & Prolonged sitting should be considered a subcategory of sedentary behaviour & 4 – Agree & Moderate Agreement & 75–89\% 4–5 & Partially Aligned \\
\addlinespace
5 & Definitions should be consistent across research, surveillance, and policy & 5 – Strongly Agree & Strong Agreement & $\ge$90\% 4–5 & Aligned \\
\bottomrule
\end{tabularx}
\end{table}

Note on Item 4: Gemini agreed that prolonged sitting is part of sedentary behaviour, but did not emphasize its subcategorical role as explicitly as the Delphi panel. This reflects some ambiguity in pre-2017 literature where terminology was still evolving, consistent with the panel’s goal of achieving clearer operational definitions.
\newpage

\section{Guiding principles for endurance running}
\label{app:app4}

\subsection{The Foundational Architecture of Endurance Adaptation}

\begin{itemize}
    \item Long-Term Consistency: Sustained, regular, and thoughtfully planned training over extended periods is the primary catalyst for profound endurance adaptations.
    \item Progressive Overload: To enhance physiological capacity, training stress must be incrementally increased (via volume, intensity, or frequency), with progression managed gradually and meticulously to ensure it remains a positive stimulus.
    \item Gradualism in Progression: Implement conservative increases in training load (e.g., 10-15\% weekly for long run duration), often with interspersed consolidation or "down" weeks to facilitate adaptation and mitigate injury risk.
    \item Individualization: Effective training programs must be bespoke frameworks tailored to the individual's unique genetic predispositions, training history, physiological characteristics, lifestyle factors, recovery patterns, and goals.
    \item Training Must Fit Life: The athlete's personal context and capacity should dictate training structure, not the other way around.
    \item Recovery and Adaptation as Critical Processes: Physiological improvements occur during periods of rest and recovery following catabolic training stress; insufficient recovery negates training benefits and increases risks.
    \item Individualized Rest Day Prescription: Decisions regarding the frequency and nature of 'true rest' (no running) days must be deeply individualized based on training load, life stress, recovery capacity, and subjective athlete feedback, as no single universal rule for rest day prescription achieved expert consensus.
    \item Primacy of Subjective Feedback in Recovery Monitoring: Attentive listening to the athlete's subjective feedback on fatigue, mood, and sleep quality, augmented by objective performance markers, is a critical coaching competency for monitoring recovery.
    \item Specificity: Training adaptations are highly specific to the nature of the stimulus applied; training must incorporate elements that replicate the demands of the target event (e.g., duration, pace, fueling, environment, terrain).
    \item Reversibility (Detraining): Fitness gains are transient and require consistent engagement with training, even during transitional phases, to preserve adaptations.
\end{itemize}

\subsection{Architecting the Training Regimen: Weekly Rhythms and Cyclical Progressions}
\begin{itemize}
    \item Polarized/Pyramidal Intensity Distribution: Effective weekly structures typically involve a high volume of low-intensity training (approx. 80\%) punctuated by a smaller proportion of targeted high-intensity sessions (approx. 20\%).
    \item Adequate Recovery Between Hard Sessions: Schedule 48 to 72 hours of recovery between high-intensity sessions to allow for physiological repair and optimal performance in subsequent quality workouts.
    \item Adaptation of Session Density by Experience: Novice runners generally benefit from fewer high-intensity days per week compared to more experienced and resilient athletes.
    \item Cautious Use of Back-to-Back Hard Days: This strategy is generally reserved for highly advanced athletes or specific, short-term blocks and requires extreme caution due to elevated overtraining risk.
    \item The Quintessential Long Run: The long run is a crucial element for extending aerobic endurance, enhancing fatigue resistance, improving fuel utilization, building musculoskeletal resilience, and practicing race-day strategies.
    \item Gradual Long Run Progression: Increase long run duration or distance gradually, adhering to progressive overload, often with weekly increases of no more than 10-15\% or by using build/consolidation week patterns.
    \item Individualized Long Run Caps: Maximum long run durations (e.g., ~3 hours or 20-22 miles for marathons) should be individualized to prevent excessive recovery demands relative to the athlete's overall training and goals.
    \item Specialized Long Runs for Ultras: Ultramarathon preparation may involve even longer single efforts or "back-to-back" long runs to simulate event demands.
    \item Conditional Inclusion of Intensity in Long Runs: Incorporating segments at goal race pace, "fast finishes," or structured pace variations can enhance race-specific fitness for experienced runners, guided by the principle of specificity.
    \item Individualized Long Run Scheduling: The specific day for the long run should be flexible, primarily determined by the individual athlete's life context, preferences, and recovery needs, as no single set of systematic rules for altering scheduling by runner type or experience achieved expert consensus.
    \item Limited Utility of "Running Doubles" for Most Recreational Runners: Performing two running sessions in a single day is generally unnecessary and potentially counterproductive for most recreational runners; focus should be on quality single sessions.
    \item Strategic Use of "Running Doubles" for Advanced Athletes: Doubles can be a tool for elite or very high-volume sub-elite athletes to accumulate mileage or incorporate advanced protocols.
    \item Beneficial "Doubling" with Run + Strength/Cross-Training: Pairing a run with strength training or low-impact cross-training on the same day is a widely beneficial practice for all levels to consolidate stress and preserve recovery days.
    \item Flexible Periodization for Recreational Runners: While periodization should ensure logical progression from general fitness to race-specific preparedness, its application for recreational runners often requires a flexible, non-linear, or "undulating" approach to accommodate life commitments.
\end{itemize}

\subsection{Calibrating Effort: The Science and Art of Training Intensity}
\begin{itemize}

    \item Multifaceted Approach to Defining Intensity Zones: Employ a range of methodologies including RPE, pace-based zones, and heart rate zones to define and communicate training intensity.
    \item Primacy of RPE/Talk Test: Rate of Perceived Exertion (RPE) and the "talk test" are highly valued for cultivating an athlete's internal sense of effort and are often the ultimate arbiters of appropriate intensity, especially for novices or when external factors affect other metrics.
    \item Utility of Pace-Based Zones: Pace zones offer objective targets useful for goal-oriented runners and specific workout structures.
    \item Cautious \& Complementary Use of Heart Rate Zones: Heart rate is a useful physiological guide but should be used as a complementary tool or governor, acknowledging its liability due to various internal and external factors.
    \item Selective Use of Advanced Intensity Monitoring: Lactate threshold testing and running power meters are generally employed by higher-level athletes or those with specialized resources.
    \item Adaptable Number of Training Zones: Typically use a 3-zone model (beneficial for novices or emphasizing polarized concepts) or a more granular 5-zone system (common and effective for many), with the choice depending on athlete experience, understanding, and the need for clarity over complexity.
    \item Flexible Adherence to Intensity Zones: Prescribed intensity zones serve as valuable guidelines, but rigid adherence can be counterproductive; RPE and somatic feedback should play a critical role, especially for low-intensity sessions.
    \item Precision for Key High-Intensity Workouts: Greater precision in hitting target efforts is generally expected for key high-intensity workouts, though minor adjustments based on daily feel are warranted.
    \item Athlete Empowerment for Intensity Adjustments: Effective coaching involves educating athletes on the purpose of different intensities and empowering them to make intelligent adjustments.
\end{itemize}

\subsection{The Runner's Toolkit: Principles of Specific Run Types (Derived from their purpose, execution, benefits, and integration)}

\begin{itemize}
    \item Recovery Runs - Principle: Perform at very low intensity for short durations primarily to promote active recovery and freshness, not direct fitness gains; complete rest or light cross-training may be more appropriate for some.
    \item Easy \& LSD Runs - Principle: Constitute the bulk of training volume at a comfortable, conversational pace to build a robust aerobic base, enhance endurance, and improve physiological efficiency (mitochondria, capillarization, fuel utilization).
    \item Tempo Runs - Principle: Sustain efforts at or near lactate threshold ("comfortably hard") to improve lactate clearance, enhance metabolic efficiency at race-relevant intensities, and build speed endurance.
    \item Lactate Threshold (LT) Intervals - Principle: Accumulate significant time at or slightly above LT intensity through repeated bouts with short recoveries to potently stimulate lactate clearance and improve sustained speed.
    \item VO2 Max Intervals - Principle: Utilize high-intensity intervals (2-5 minutes at ~3k-5k pace) with comparable recovery periods to directly challenge and improve maximal oxygen uptake and aerobic power.
    \item Strides - Principle: Incorporate short, controlled accelerations (not all-out sprints) regularly to improve neuromuscular coordination, running mechanics at faster speeds, and maintain leg turnover with minimal metabolic stress.
\end{itemize}

\subsection{Strategic Cross-Training: Enhancing Fitness and Mitigating Risk}
\begin{itemize}
    \item Cross-Training as a Strategic Supplement: Engage in non-running exercise modalities to enhance overall fitness, reduce impact stress, aid injury prevention/management, facilitate recovery, and provide psychological variety.
    \item Prioritize Low/Non-Impact Modalities for Recovery/Injury Management: Utilize activities like cycling, swimming, aqua jogging, or elliptical training to maintain cardiovascular fitness while minimizing impact.
    \item Aqua Jogging for Run-Specific Fitness Maintenance: Deep water running effectively maintains run-specific fitness and neuromuscular patterns during periods of non-running.
    \item Balance Cross-Training with Running Specificity: While beneficial, cross-training should supplement, not entirely replace, running for a healthy athlete, especially as a key race approaches, to adhere to the principle of specificity.
\end{itemize}

\subsection{Synergistic Elements: Enhancing Performance and Resilience}

\begin{itemize}
    \item Integral Role of Strength Training: Implement 1-2 weekly sessions of functional, runner-specific strength training year-round to prevent injuries, improve running economy, and build musculoskeletal resilience.
    \item Periodize and Integrate Strength Training: Shift the focus of strength training (e.g., foundational strength in base phase, maintenance or power-endurance in race phase) and integrate it to complement running, often on harder running days or days allowing recovery.
    \item Intra-Run Fueling for Prolonged Efforts: For efforts exceeding 75-90 minutes, consume carbohydrates (e.g., 30-60g/hr, increasing to 60-90g/hr+ for longer events if tolerated) to maintain blood glucose and spare glycogen.
    \item Individualized "Gut Training" for Fueling: Emphasize individual experimentation with fuel types, amounts, and timing during training to adapt the digestive system and develop a personalized, effective race-day strategy. (This reflects the Conditional Consensus on varying fueling guidance).
    \item Strategic Tapering for Peak Performance: Implement a taper (1-3 weeks, varying by race distance) by significantly reducing training volume while largely maintaining intensity and frequency to shed fatigue and optimize race readiness.
\end{itemize}

\subsection{The Shadow of Excessive Load: Understanding Overtraining Syndrome}

\begin{itemize}
    \item Distinguish FOR, NFOR, and OTS: Understand the spectrum of training maladaptation from functional overreaching (planned, beneficial) to non-functional overreaching (prolonged fatigue) and Overtraining Syndrome (severe, long-lasting).
    \item Multifactorial Etiology of Overtraining: Recognize that NFOR/OTS stem from chronic excessive total stress (training errors, insufficient recovery, nutritional deficiencies like RED-S, life stress) relative to coping capacity.
    \item Prioritize Prevention of Overtraining: Focus on individualized and periodized training, gradual progressive overload, systematic monitoring, prioritizing recovery, optimizing nutrition (maintaining energy availability), managing life stress, and athlete/coach education.
    \item Systematic Monitoring - Subjective Feedback as Cornerstone: Prioritize subjective athlete feedback (mood, fatigue, sleep, RPE) and performance cues for monitoring training load and recovery.
    \item Systematic Monitoring - Caution and Individualization with Technology: Approach wearable technology metrics (HRV, readiness scores) with caution as supplementary tools, acknowledging varied expert views on their standalone reliability and the need for individualized interpretation alongside subjective feedback. No single technological monitoring strategy or systematic differentiation by athlete type achieved universal expert consensus.
    \item Systematic Monitoring - Response to Early Fatigue: Recognize that expert approaches to managing sessions missed due to early fatigue vary (e.g., structure preservation vs. immediate individualization based on context); no single rule applies, requiring careful coaching judgment.
    \item Rest as Cornerstone of NFOR/OTS Management: Prolonged rest is the primary treatment for NFOR/OTS, with a very gradual, monitored return to training only after full recovery and addressing underlying contributors (e.g., LEA/RED-S).
\end{itemize}

\subsection{The Human Element: Coaching Philosophy and the Athlete-Coach Dyad}
\begin{itemize}

    \item Athlete-Centered Coaching Philosophy: Prioritize the athlete's long-term health, safety, and well-being above immediate performance, fostering sustainable engagement, enjoyment, intrinsic motivation, and autonomy.
    \item Transparent and Honest Communication: Build coaching relationships on transparency, honesty, realistic expectations, and evidence-based practices adapted to individual contexts.
    \item Individualized and Multifaceted Communication: Tailor communication modes and frequency to the athlete's personality and needs. However, recognize that expert philosophies diverge on whether communication strategies should be systematically and formally adjusted based on broad runner categories (novice, intermediate, sub-elite), with the unifying principle being effective, empathetic, and responsive communication with the specific individual.
    \item Critical and Supplementary Use of Digital Tools: Utilize training platforms for planning/feedback and wearable technology as supplementary aids, but always prioritize coaching judgment and athlete subjective experience over raw data. Acknowledge varied expert views on the reliability of some wearable data when integrating it.
    \item Dynamic and Responsive Plan Modification: Drive training plan adjustments primarily by athlete subjective feedback, performance markers, and life stressors, balancing planned structure with real-time responsiveness.
    \item Cultivate Athlete Self-Awareness: Foster athlete autonomy in understanding their bodies and making sensible minor training adjustments as experience grows.
    \item Uphold Strong Professional Values: Commit to athlete safety, promote positive body image and mental health, foster joyful sport engagement, and recognize limits of expertise by referring when necessary.
\end{itemize}
\newpage
\section{Guiding principles for resistance and mixed cardio/strength training}
\label{app:app5}

The expert consensus process yielded a comprehensive framework of principles guiding the application of resistance and mixed-modality (concurrent) training. The key dimensions of this framework are summarized below:

\begin{itemize}
    \item Training Variables and Goal Alignment: A foundational outcome was the articulation of how training volume (primarily defined as effective working sets, but acknowledging total repetitions and volume load), intensity (load, often guided by \%1RM, RPE, or RIR), and effort (proximity to technical failure) must be synergistically managed and meticulously aligned with primary training goals (whether strength, hypertrophy, endurance, or a combination in mixed-modality approaches). Specific guidelines were established for optimizing these variables for hypertrophy, strength development (including strength without mass), muscular endurance, metabolic conditioning, and muscle preservation during fat-loss phases. The critical role of recoverable volume ceilings and planned “deloads” (both proactive and reactive) for sustainable progress was consistently emphasized.
    \item Progression Strategies: The consensus highlighted an evolution in progression strategies, from simple linear and double-progression models suitable for novices (prioritizing technical mastery and confidence with conservative initial loads) to more sophisticated periodization models (e.g., ‘block’, undulating – “DUP”/“WUP”) for intermediate and advanced individuals engaging in resistance or mixed-modality training. The development of autoregulation was identified as a core skill, enabling performance-driven adjustments to maintain an optimal training stimulus relative to daily readiness.
    \item Application of Training to Failure: A nuanced perspective on training to failure emerged, positioning it as a specific tool rather than a default method. Emphasis was placed on differentiating technical failure (form breakdown, the preferred stopping point for most training) from muscular failure. While selective, cautious use of muscular failure was deemed potentially beneficial for advanced trainees in specific hypertrophy or peaking contexts (e.g., final sets of isolation exercises, supervised AMRAPs – As Many Reps As Possible), consistent training near, but not to, failure (e.g., 1-3 RIR) was recognized as yielding similar benefits with less fatigue and risk, particularly for compound lifts and special populations.
    \item Exercise Order and Session Structure: The principles affirmed that exercise order should prioritize skill, safety, and the primary adaptation goal. Typically, complex, high-load, multi-joint movements precede simpler, lower-load, or isolation exercises. However, context-specific adjustments, such as pre-activation drills or interleaved exercises in metabolic circuits (common in mixed-modality training), are permissible when aligned with the session's purpose and do not compromise safety or key performance outcomes.
    \item Concurrent Training Integration: For integrating strength and endurance (mixed cardio/strength training), the consensus stressed strategic scheduling to mitigate interference. This includes prioritizing the primary goal in session sequencing (e.g., resistance training before cardio if strength is key), selecting lower-impact cardio modalities (e.g., cycling, incline walking), managing cardio intensity (especially HIIT) around demanding resistance sessions, and ensuring adequate recovery between conflicting stimuli.
    \item Recovery as a Foundational Component: Recovery was framed not as passive rest alone, but as an active, planned, and individualized component of training architecture, crucial for both resistance and mixed-modality programs. This includes structured deloads, attention to sleep quality and quantity, nutritional support, stress management, and the utility of low-intensity active recovery methods like walking. The need to tailor recovery strategies to individual profiles (e.g., age, occupational stress, training status, total training load from combined cardio and strength work) was a strong theme.
    \item Autoregulation as an Essential Skill: Autoregulation was identified as a crucial skill for long-term, sustainable training, enabling individuals to modulate training variables based on daily readiness in both resistance-focused and mixed-modality regimens. The principles support introducing basic autoregulatory concepts (like RIR) to novices and developing more refined RPE-based and performance-based adjustments in experienced trainees.
    \item Multi-faceted Feedback Interpretation: Effective program modification relies on synthesizing a composite of subjective feedback (e.g., mood, soreness, RPE, motivation) and objective indicators (e.g., performance data, wearable metrics where appropriate and validated). No single metric is definitive; rather, a dialogue between data streams, contextualized by individual trends, should guide adjustments for any training type.
    \item Adaptations for Special Populations: The framework strongly advocates for structurally adapted programming for special populations (e.g., older adults, post-rehab, perimenopausal individuals, those with specific health conditions or anxieties), emphasizing that adaptation should facilitate safe participation and progress towards ambitious goals in resistance and mixed cardio/strength training, rather than merely reducing expectations. This involves modified loading, exercise selection, tempo, and a focus on building confidence and self-efficacy.
    \item Dynamic and Evolving Programs: Finally, the consensus underscored that training programs must be dynamic systems that evolve with the individual, not static templates, whether for pure resistance training or mixed-modality approaches. Regular review and revision based on ongoing feedback, changing goals, and life circumstances are essential for sustained adherence and progress. In summary, the consensus process produced a rich, multi-layered set of principles that emphasize individualization, evidence-based practice, and a holistic, adaptive approach to resistance and mixed-modality training.
\end{itemize}

\end{document}